\documentclass[10pt,twocolumn,letterpaper]{article}

\usepackage{iccv}
\usepackage{times}
\usepackage{epsfig}
\usepackage{graphicx}
\usepackage{amsmath}
\usepackage{amssymb}
\usepackage{tabularx}
\usepackage{theorem}
\usepackage[table]{xcolor}

\usepackage[pagebackref=true,breaklinks=true,letterpaper=true,colorlinks,bookmarks=false]{hyperref}

\iccvfinalcopy 

\def\iccvPaperID{1574} 

\ificcvfinal\fi

\newtheorem{theorem}{Theorem}

\newcommand{\X}{\mathcal{X}}
\newcommand{\Y}{\mathcal{Y}}
\newcommand{\LL}{\mathcal{L}}

\begin{document}
	
	\title{Learning in an Uncertain World:\\Representing Ambiguity Through Multiple Hypotheses}
	
	\author{Christian Rupprecht$^{1,2}$ \and Iro Laina$^{1}$ \and Robert DiPietro$^{2}$ \and Maximilian Baust$^{1}$ \and Federico Tombari$^{1}$ \and Nassir Navab$^{1,2}$ \and Gregory D. Hager$^{2}$\\ \\ \and $^1$Technische Universit\"at M\"unchen, Munich, Germany \and $^2$Johns Hopkins University, Baltimore MD, USA \and \color{gray} International Conference on Computer Vision (ICCV) 2017
	}
	
	\maketitle

	\begin{abstract}
		Many prediction tasks contain uncertainty.
		In some cases, uncertainty is inherent in the task itself. In future prediction, for example, many distinct outcomes are equally valid.
		In other cases, uncertainty arises from the way data is labeled.
		For example, in object detection, many objects of interest often go unlabeled, and in human pose estimation, occluded joints are often labeled with ambiguous values.
		In this work we focus on a principled approach for handling such scenarios.
		In particular, we propose a framework for reformulating existing single-prediction models as multiple hypothesis prediction (MHP) models and an associated meta loss and optimization procedure to train them.
		To demonstrate our approach, we consider four diverse applications: human pose estimation, future prediction, image classification and segmentation.
		We find that MHP models outperform their single-hypothesis counterparts in all cases, and that MHP models simultaneously expose valuable insights into the variability of predictions.
	\end{abstract}
	
	\section{Introduction}
Dealing with uncertainty is fundamental in many tasks.
Given an image, for example, one might think \textit{this is either an alpaca or a llama, but it is certainly not an elephant}.
When predicting the behavior of other drivers on the road, we also tend to make good guesses based on our learned expectations. If someone is driving forward in the right lane, one might think \textit{they will probably continue straight or take a right turn soon}.
In addition, uncertainty models incomplete information.
For example, we may not be able to distinguish a mug from a cup if its handle is not visible. 
In short, when confronted with a situation that we are not sure about, we tend to produce multiple plausible hypotheses.



In this work, we present a framework for multiple hypothesis prediction (MHP) which extends traditional single-loss, single-output systems to multiple outputs and which provides a piece-wise constant approximation of the conditional output space.
To achieve this, we propose a probabilistic formulation and show that minimizing this formulation yields a Voronoi tessellation in the output space that is induced by the chosen loss.
Furthermore, we explain how this theoretical framework can be used in practice to train Convolutional Neural Networks (CNNs) that predict multiple hypotheses.
By employing a novel \emph{meta loss}, training can be achieved through standard procedures, i.e. gradient descent and backpropagation.

Our framework has the following benefits. First, it is general in the sense that it can easily retrofit any CNN architecture and loss function or even other learning methods, thus enabling multiple predictions for a wide variety of tasks. Second, it exposes the variance of different hypotheses, thus providing insights into our model and predictions. Third, as shown in our experiments, allowing multiple hypotheses often improves performance. For example, in the case of regression, single hypothesis prediction (SHP) models often average over distinct modes, thus resulting in unrealistic, blurred predictions. MHP models are capable of overcoming this issue, as demonstrated in Figure \ref{fig:intersection}.

In an extensive experimental evaluation, we consider four applications of our model: 
human pose estimation, future frame prediction, multi-label classification and semantic segmentation. Despite their vastly different nature, all four tasks show that MHP models improve over their corresponding SHP models and also provide additional insights into the model and into prediction variability.

We proceed in the next section by describing the related work. In Section \ref{sec:methods}, we describe our approach and detail the theory of the proposed multiple prediction framework. Next, in Section \ref{sec:experiments}, we describe our experiments; here, we solidify the ideas from Section \ref{sec:methods} and demonstrate the benefits of MHP models. Finally, in Section \ref{conclusions}, we conclude.

	\section{Related Work}
\label{sec:relatedWork}
%
CNNs \cite{lecun1998gradient} have been shown to be flexible function approximators and have been used extensively for a wide variety of tasks, such as image classification \cite{krizhevsky2012imagenet, he2015deep}, object detection \cite{renNIPS15fasterrcnn} and semantic segmentation \cite{chen14semantic}.
However, the problem of predicting multiple hypotheses in computer vision has been addressed less extensively in the literature and often under different names and assumptions.

Mixture density networks (MDNs) \cite{bishop1994mixture} are neural networks which learn the parameters of a Gaussian mixture model to deal with multimodal regression tasks. MDNs differ from our approach in two major ways. First, MDNs are limited to regression, whereas MHP models are loss agnostic and therefore extend naturally to many tasks. Second, rather than predicting a mixture of Gaussians as in MDNs, MHP models yield a Voronoi tessellation in the output space which is induced by the chosen loss. In our experiments (Section \ref{sec:experiments}) we also show that MDNs can be difficult to train in higher dimensions due to numerical instabilities in high dimensional, multivariate Gaussian distributions.

Multiple Choice Learning \cite{dey2015predicting,lee2017confident,lee2016stochastic} is a line of work that focuses on predicting multiple possibilities for each input, while in \cite{guzman2014efficiently} the goal is to also enforce diversity among the predictions. In closely related work, Lee \etal~\cite{lee2016stochastic}, train an ensemble of networks with a minimum formulation that is similar to ours. We extend these ideas by providing a mathematical understanding why this formulation is beneficial, extend to regression tasks and introduce a relaxation that helps convergence. Instead of training separate networks for each choice, we use a shared architecture for the hypotheses which saves a considerable amount of parameters and enables information exchange between predictions.

Gao \etal~\cite{gao2016deep} deal with label ambiguity in different domains, such as age estimation and image classification, and study the improvement on performance when training CNNs with soft, probabilistic class assignments and Kullback-Leibler (KL) divergence. 
Geng \etal~also propose multi-label approaches for age estimation \cite{geng2013facial} and head pose estimation \cite{geng2014head}. 

Unlike single-label image classification, multi-label recognition is more general and relevant in real applications, as objects usually appear in their natural environment along with more objects of different categories. 
This direction is receiving increasing attention as many approaches have been proposed to handle the label ambiguity in image classification.
Wang \etal~\cite{wang2016cnn} propose to model label dependency by using a recurrent neural network (RNN) on top of a CNN.
This task has also been tackled using deep convolutional ranking \cite{gong2013deep}. Several other works propose pipelines of object proposals or use ground truth bounding boxes and/or classifiers to predict multiple labels~\cite{wang2016beyond, wei2014cnn, yang2015can}.

In future prediction, uncertainty is inherent in the task itself. 
Especially for robotic applications, it is sometimes crucial to predict what humans will be doing \cite{koppula2016anticipating}. 
In \cite{yuen2010data} Yuen and Torralba transfer motion from a video database to images. 
Lerer \etal~\cite{lerer2016learning} predict the configuration and fall probability of block towers.
Multiple predictions have also been used by Vondrick \etal~\cite{vondrick2015anticipating} for future frame anticipation. 
In \cite{fouhey2014predicting} Fouhey and Zitnick predict spatio-temporal likelihood distributions for humans in cartoons and pictures. Walker \etal\cite{walker2016uncertain} deal with uncertainty by predicting dense trajectories of motion using a variational autoencoder.

Except \cite{bishop1994mixture} and possibly \cite{gao2016deep} that addresses classification, all these works are driven by a specific application, rendering their translation to other tasks not straightforward.

There also exists some work that focuses on obtaining confidences for the predictions from the network.
Gal \etal\cite{gal2015dropout} instead analyze how sampling from dropout layers can be used to extract uncertainty estimates from the network.
Kingma \etal\cite{kingma2013auto} propose a stochastic gradient variational Bayes estimator to estimate the posterior probability.

As our method is based on the mathematical concept of (centroidal) Voronoi tesselations, we refer the interested reader to the more general book of Okabe \etal~\cite{okabe2009spatial} or to Du \etal~\cite{du1999centroidal}, which is more closely related to this work.
However, detailed knowledge of Voronoi tesselations is not necessary to understand our approach.
	\section{Methods}
\label{sec:methods}
Here, we describe the proposed ambiguity-aware model and investigate its relationship to traditional (unambiguous) prediction models.
We represent the vector space of input variables by $\X$ and the vector space of output variables or \textit{labels} by $\Y.$ We assume that we are given a set of $N$ training tuples $(x_i, y_i)$, where $i = 1,\ldots,N$.
Furthermore, we denote the joint probability density over input variables and labels by $p(x,y) = p(y | x) p(x)$, 
where $p(y|x)$ denotes the conditional probability for the label $y$ given the input $x$.
\subsection{The Unambiguous Prediction Model}
In a supervised learning scenario, we are interested in training a predictor $f_\theta:\X\rightarrow\Y$, parameterized by $\theta\in\mathbb{R}^n$, such that the expected error
\begin{equation}
\frac{1}{N}\sum_{i=1}^N \LL(f_\theta(x_i), y_i)
\label{eqn:unambiguous_discrete}
\end{equation}
is minimized, where it is assumed that the training samples follow
$p(x,y)$. Here, $\LL$ can be any loss function, for example the classical $\ell_2$-loss 
\begin{equation}
	\LL_2(u,v) = \frac{1}{2}||u-v||^2_2.
\end{equation}
For sufficiently large $N$, Equation \eqref{eqn:unambiguous_discrete} yields a good approximation of the continuous formulation 
\begin{equation}
\int_\X \int_Y \LL(f_\theta(x), y) p(x,y)\;dy\;dx.
\label{eqn:unambiguous}
\end{equation}
In that case, Equation \eqref{eqn:unambiguous} is minimized by the conditional average (see e.g.~\cite{kolmogorov1950foundations}).
\begin{equation}
f_{\theta}(x) = \int_\Y y\cdot p(y|x)\;dy.
\label{eqn:conavg}
\end{equation}
However, depending on the complexity of the conditional density $p(y|x)$, the conditional average can be a poor representation.
For example, in a mixture model of two well separated Gaussian distributions, the expected value falls between the two means, where the probability density is low.
\subsection{The Ambiguous Prediction Model}
If, given $x$, single predictions essentially represent the expected value distribution with a single constant value $f_{\theta}(x)$, then it follows that multiple values might serve as a better approximation. 
To this end, let us assume that we develop a prediction function that is capable of providing $M$ predictions:
\begin{equation}
f_\theta(x) = (f^1_\theta(x),\ldots, f^M_\theta(x)).
\end{equation}
The idea is, to compute the loss $\LL$ always for the closest of the $M$ predictions.
Instead of \eqref{eqn:unambiguous}, we propose to minimize
\begin{equation}
\int_\X \sum_{j=1}^M \int_{\Y_j(x)} \LL(f^j_\theta(x), y) p(x,y)\;dy\;dx,
\label{eqn:ambiguous}
\end{equation}
where we consider the Voronoi tessellation of the label space $\Y = \cup_{i=1}^M\overline{\Y_i}$ which is induced by $M$ generators $g^j(x)$ and the loss $\LL$:
\begin{equation}
\Y_j(x) = \left\{y\in \Y : \LL(g^j(x),y)<\LL(g^k(x),y)\; \forall k\neq j \right\}.
\label{eqn:tesselation}
\end{equation}
Intuitively, the Voronoi tessellation follows the idea that each cell contains all points that are closest to its generator. Here, the closeness is defined by the loss $\LL$. Thus, (\ref{eqn:ambiguous}) divides the space into $M$ Voronoi cells generated by the predicted hypotheses $f^j_\theta(x)$ and aggregates the loss from each.

In a typical regression case $\LL$ is chosen as the classical $\ell_2$-loss. In that case, the loss directly translates to intuitive geometric understanding of distance in the output space. For this case, we can further show an interesting property that helps understanding the method. If the density $p(x,y)$ satisfies mild regularity conditions (\ie it vanishes only on a subset of measure zero), the following proposition holds.
\begin{theorem}[Minimizer of \ref{eqn:ambiguous}]
	A necessary condition for Equation \eqref{eqn:ambiguous} to be minimal is that the generators $g^j(x)$ are identical to the predictors $f^j_\theta(x),$ and both correspond to a centroidal Voronoi tesselation:
	\begin{equation}\label{eq:voronoi}
	g^j(x) = f^j_\theta(x) = \frac{\int_{\Y_j} \LL(f^j_\theta(x), y) p(y|x)\;dy}{\int_{\Y_j} p(y|x)\;dy},
	\end{equation} 
	i.e. $f^j_\theta$ predicts the conditional mean of the Voronoi cell it defines.
\end{theorem}
\textit{Proof.} At first we note that Equation \eqref{eqn:ambiguous} can be minimized in a point-wise fashion w.r.t.~$x$ as both $\LL$ and $p(x,y)$ are non-negative.
Thus, it suffices to minimize 
\begin{equation}
\sum_{j=1}^M \int_{\Y_j(x)} \LL(f^j_\theta(x), y) p(x,y)\;dy
\label{eq:point-wise}
\end{equation}
for every $x\in \X$.
The second equality in Equation \eqref{eq:voronoi} follows by computing the first variation w.r.t.~$f^j_\theta$ as done in \cite[Proposition 3.1]{du1999centroidal}:
\begin{equation}
f^j_\theta(x) = \frac{\int_{\Y_j} \LL(f^j_\theta(x), y) p(x,y)\;dy}{\int_{\Y_j} p(x,y)\;dy}.
\label{eq:intermediate}
\end{equation}
Using the factorization $p(x,y) = p(y | x) p(x)$ and noting that the integration does not depend on $x$, we pull $p(x)$ out of the integrals and eventually replace $p(x,y)$ by $p(y|x)$ in Equation \eqref{eq:intermediate}.

The first equality in Equation \eqref{eq:voronoi}  can be proven by contradiction: 
If the generators $g^j(x)$ do not coincide with $f^j_\theta(x)$, it is possible to find subsets of $\Y$ which have non-vanishing measure and where Equation \eqref{eq:point-wise} cannot be minimal.
For a more detailed derivation, we refer to \cite{du1999centroidal}. Intuitively, minimizing Equation \eqref{eqn:ambiguous} corresponds to finding an optimal piecewise constant approximation of the conditional distribution of labels in the output space. The hypotheses will tessellate the space into cells with minimal expected loss to their conditional average (see Equation \ref{eqn:conavg}).
\subsection{Minimization Scheme}\label{sec:minimizationscheme}
In this section, we detail how to compute $f^j_\theta$ from a set of examples $(x_i,y_i), i \in \{1,\ldots,N\}$. 
Due to their flexibility and success as general function approximators we choose to model $f^j_\theta$ with a (deep) neural network, more specifically a CNN, since our input domain $\X$ will later be images. 
It is important to note, however, that the general formulation of the energy in Equation \eqref{eqn:ambiguous} leaves the choice of $f^j_\theta$ free and any machine learning model could potentially be used. 

To minimize Equation \eqref{eqn:ambiguous} we propose an algorithm for training neural networks with back-propagation.
Our minimization scheme can be summarized in five steps:
\begin{enumerate}
	\item Create the set of $M$ generators $f^j_\theta(x_i), j\in\{1,\ldots M\}$  for each training sample $(x_i,y_i)$ by a forward pass though the network.
	\item Build the tessellation  $\Y_j(x_i)$ of $\Y$ using the generators $f^j_\theta(x_i)$, Equation \eqref{eqn:tesselation} and a loss function $\LL$.
	\item Compute gradients for each Voronoi cell $\frac{\partial}{\partial \theta}\frac{1}{|\Y_j|}\sum_{y_i\in\Y_j}\LL(f^j_\theta(x_i), y_i),$
	where $|\Y_j|$ denotes the cardinality of $\Y_i$.
	\item Perform an update step of $f^j_\theta(x_i)$ using the gradients per hypothesis $j$ from the previous step.
	\item If a convergence criterion is fulfilled: terminate. Otherwise continue with step 1.
\end{enumerate}
This algorithm can easily be implemented using a meta-loss $\mathcal{M}$ based on Equation \eqref{eqn:ambiguous}. We call $\mathcal{M}$ a meta loss because it operates on top of any given standard loss $\mathcal{L}$:
\begin{equation}\label{eq:multiloss}
\mathcal{M}(f_\theta(x_i),y_i) = \sum_{j=1}^{M}\delta(y_i \in \Y_j(x_i))\mathcal{L}(f^j_\theta(x_i),y_i).
\end{equation}
We use the Kronecker delta $\delta$ that returns $1$ when its condition is true and $0$ otherwise, in order to select the best hypothesis $f^j_\theta(x_i)$ for a given label $y_i$. This algorithm can be seen as an extension of Lloyd's Method \cite{lloyd1982least} to gradient descent methods used for training with back-propagation.

One simple way to transform an existing network into a MHP model is to replicate the output layer $M$ times (with different initializations). During training, each of these $M$ predictions is compared to the ground truth label based on the original loss metric but weighted by $\delta$ as the meta loss suggests (Equation \eqref{eq:multiloss}). Similarly, during back-propagation, $\delta$ provides a weight for the resulting gradients of the hypotheses. 
This algorithm can also be seen as a type of Expectation Maximization (EM) method. In the E-step, the association of the true label $y_i$ to a prediction $f^j_\theta(x_i)$ is computed and in the M-step the parameters of the predictor are updated to better predict the target $y_i$ in label space.

In practice, we have to relax $\delta$ to be able to minimize $\mathcal{M}$ with stochastic gradient descent. 
The problem comes from the fact that the generators $f^j_\theta(x)$ may be initialized so far from the target labels $y$ that all $y$ lie in a single Voronoi cell $k$. 
In that case only the $k$-th generator $f^k_\theta(x)$ gets updated since $\delta(y_i \in \Y_j(x_i)) = 0,~\forall j\neq k$. To address this issue, we relax the hard assignment using a weight $0<\epsilon<1$:
\begin{equation}\label{eq:multiloss_weight}
\hat{\delta}(a) = 
\begin{cases} 
1-\epsilon & \text{if } a \text{ is true,} \\
\frac{\epsilon}{M-1} & \text{else.}
\end{cases}
\end{equation}
A label $y$ is now assigned to the closest hypothesis $f^k_\theta(x)$ with a weight of $1-\epsilon$ and with $\frac{\epsilon}{M-1}$ to all remaining hypotheses.
This formulation ensures that $\sum_{j=1}^{M}\hat{\delta}(y_i \in \Y_j(x_i)) = 1$. 
Additionally, we adapt the concept from \cite{srivastava2014dropout} to drop out full predictions with some low probability (1\% in our experiments).
Such treatment effectively introduces some randomness in the selection of the best hypothesis, such that "weaker" predictions will not vanish during training.  
Now, even in the previously discussed case of a bad initialization, the non-selected predictions will slowly evolve until their Voronoi regions contain some training samples.

It is noteworthy that our formulation of the meta-loss $\mathcal{M}$ (see Equation \eqref{eq:multiloss}) is agnostic to the choice of loss function $\mathcal{L}$, as long as $\mathcal{L}$ is to be minimized during the learning process. We also show the generic applicability of this method in Section \ref{sec:experiments}, where we use $\mathcal{M}$ with three different loss functions $\mathcal{L}$ and four different CNN architectures for $f_\theta$.

While the number of hypotheses $M$ is a hyper-parameter for this model, we do not see any deterioration in performance when increasing $M$ in all regression problems.  
In fact, almost every method that models posterior probabilities needs some form of hand-tuned model parameter: $k$-means ($k$), MDNs~\cite{bishop1994mixture} (number of Gaussians $m$). 

	\section{Experiments}
\label{sec:experiments}
In this section, we perform extensive experiments to validate different properties of the proposed approach. 
\begin{enumerate}
	\item Using a 2D toy example, we show an intuition of the Voronoi representation of the model in Section \ref{sec:voronoiEx}.
	\item We use human pose estimation as a standard low-dimensional regression problem in Section \ref{sec:humanpose} to highlight the underlying information that can be obtained by analyzing the variance across hypotheses.
	\item In the scenario of future frame-prediction, we demonstrate that the approach generalizes to high-dimensional problems and that the predicted images become sharper with more predictions (Section \ref{sec:frameprediction}).
	\item Finally, the ability to handle discrete problems is demonstrated in Sections \ref{sec:classification} and \ref{sec:segmentation} in the context of multi-label image classification and segmentation.
\end{enumerate}

We emphasize that for all these applications we use simple, single-stage models to study the behavior and evaluate the concept of multiple predictions directly.
Complex multi-stage pipelines would benefit both SHP and MHP models and likely improve their performance, but obscure the analysis of the raw MHP framework.
Thus, we learn every task end-to-end by training or fine-tuning previously proposed CNN architectures~\cite{belagiannis2015robust,he2015deep,laina2016deeper,long2015fully}.
All experiments were performed on a single NVIDIA TitanX with 12GB GPU memory. 
It is important to note that the influence of the number of predictions $M$ on training time is usually negligible as it affects only the last layer of the network and has only an insignificant impact on the overall execution time of the architecture.
In all experiments we set the association relaxation to $\epsilon = 0.05$.
We refer to our model as $M$-MHP, denoting a network trained to predict $M$ hypotheses. The corresponding single prediction model is named as SHP.

\subsection{Temporal 2D Distribution}
\label{sec:voronoiEx}
\begin{figure}[t]
	\centering
	\includegraphics[width=0.95\linewidth]{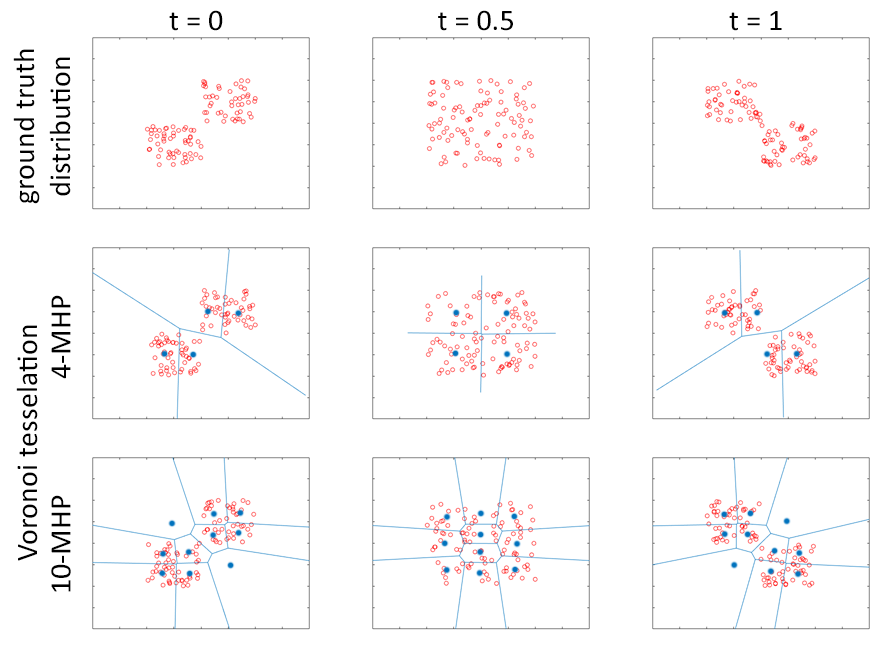}
	\caption{\textbf{Temporal 2D Distribution Illustration.} Red points are drawn from the true underlying distribution, blue points show predictions, and blue lines highlight the resulting Voronoi regions.}	
	\label{fig:2dvoronoi}
\end{figure}
We start with a toy example of a two-dimensional distribution that changes over time $t \in [0,1]$ to demonstrate the representation that is built with an MHP model.
Intuitively, we split a zero-centered square into 4 equal regions, and we smoothly transition from having high probability mass in the lower-left and top-right quadrants to having high probability mass in the upper-left and lower-right quadrants. At $t=\frac{1}{2}$ the whole square has uniform probability.
More precisely, the 2D plane is divided into five sections $S_i$:
\begin{eqnarray}
	 S_1 = &[-1, 0) \times [-1, 0) \subset \mathbb{R}^2,\\
	 S_2 = &[-1, 0) \times [0, 1] \subset \mathbb{R}^2, \\
	 S_3 = &[0, 1] \times [-1, 0) \subset \mathbb{R}^2, \\
	 S_4 = &[0, 1] \times [0, 1] \subset \mathbb{R}^2, \\
	 S_5 = &\mathbb{R}^2 \backslash  \{S_1 \cup S_2 \cup S_3 \cup S_4 \}.
\end{eqnarray}
We then create a distribution that depends on time, by first defining the probability that $S_i$ get selected as $p(S_1) = p(S_4) = \frac{1 - t}{2}$, $p(S_2) = p(S_3) = \frac{t}{2}$ and $p(S_5) = 0$. When a region is selected, a point is sampled from it uniformly. This creates the distribution that can be seen in the first row of Fig.~\ref{fig:2dvoronoi}. It transitions smoothly between the three states.

We then train a simple three-layer fully connected network with 50 neurons in both hidden layers and ReLU as activation function. The input is the time $t$ and the output hypotheses are 2D coordinates for each prediction. The network is trained with $\ell_2$-loss as objective. We then show the Voronoi tessellation for 4-MHP and 10-MHP in the bottom two rows of Figure \ref{fig:2dvoronoi}. The model is able to adapt the hypotheses to the conditional distribution and divides the space into Voronoi cells that match the regions. With more hypotheses the tessellation becomes finer. 

After having demonstrated the output representation of the model, we apply the approach to real-world problems in the following sections.

\subsection{Human Pose Estimation}
\label{sec:humanpose}
For the second experiment we move from 1D input, 2D output to image input and 24-dimensional output. 2D human pose estimation is the task of regressing the pixel locations of the joints of a human body.
In this experiment we demonstrate that our multiple prediction framework not only works with a robust loss function, but also the variation of the predictions can be used to measure the confidence of the model.
Here, we adapt the model from Belagiannis \etal~\cite{belagiannis2015robust}, which uses Tukey's bi-weight function as an objective, in order to study the behavior of another loss function $\mathcal{L}$ in the MHP setting.
To better understand the gain of increasing $M$, we evaluate the strict PCP score using an oracle to select the best hypothesis which results in SHP: 59.7\%, 2-MHP: 60.0\%, 5-MHP: 61.2\%, 10-MHP: 62.8\%. With increasing number of predictions the method is able to model the output space more and more precisely. This means that secondary approaches can be designed to select good hypotheses to further improve results. 
\begin{figure}[t]
	\centering
	\includegraphics[width=0.8\linewidth]{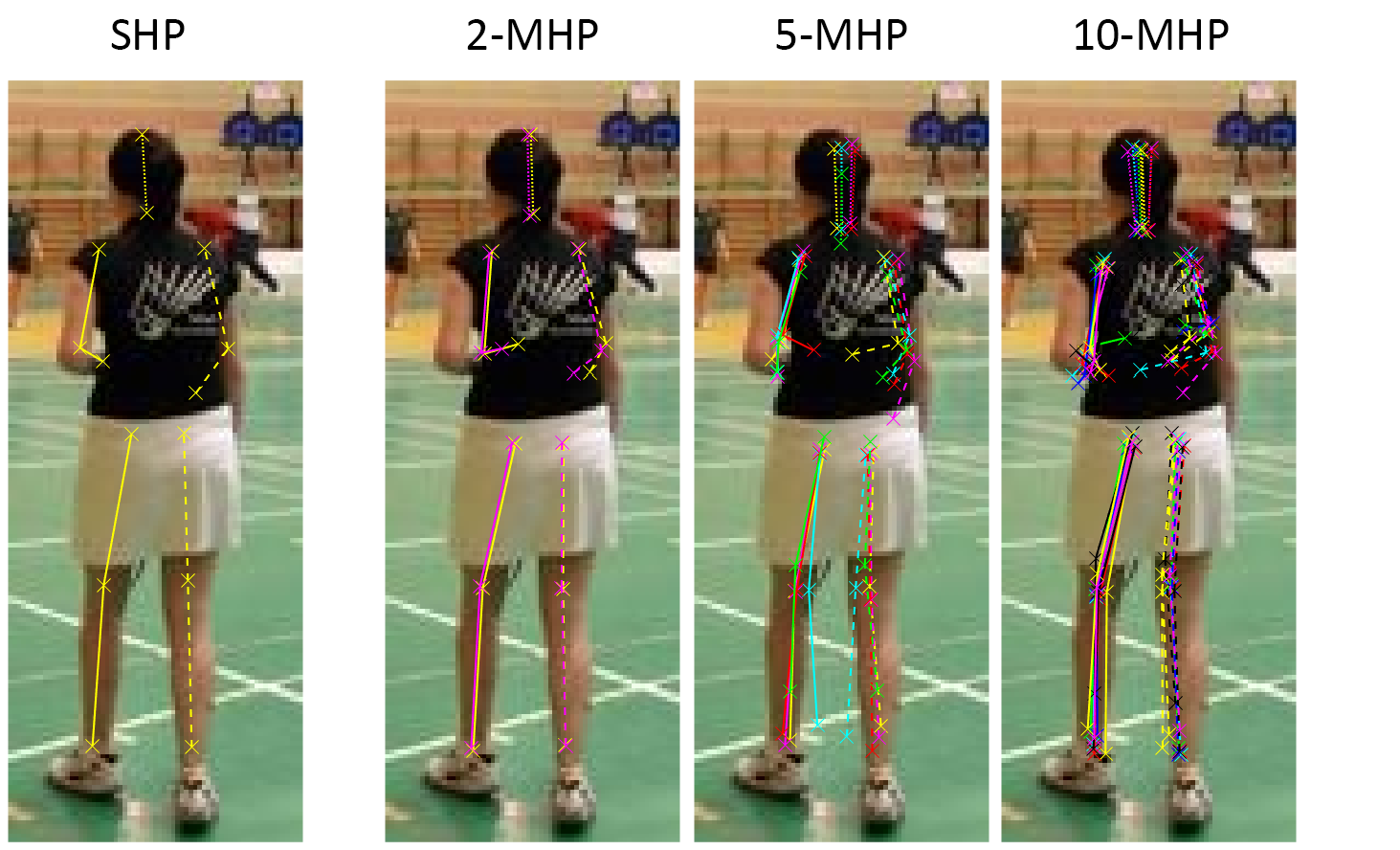}
	\caption{\textbf{Human Pose Estimation on the LSP dataset.} We show the predicted human pose for an image with SHP and with two, five, and ten MHPs. We observe the uncertainty of the hand positions in the high variance with multiple predictions. Joints like shoulders and hips are easy to detect and also vary much less.}	
	\label{fig:humanpose}
\end{figure}

Figure \ref{fig:humanpose} shows qualitative results for human pose estimation for different $M$. We can see that the variance of the predictions of the occluded joints (both wrists) is higher than the variance of directly visible joints like the shoulder or the hips.

\begin{table}[t]
	\centering
	\small
\begin{tabular}{ l|r|r|r|r|r|r }
	\textbf{body part} & ankle & knee & hip & wrist & elbow & shldr   \\
	\hline
	dist.~visible & 4.8 & 3.0 & 1.9 & 5.0 & 3.1 & 2.3 \\ 
	dist.~occl. & 5.9 & 3.7 & 2.4 & 5.1 & 3.3 & 2.6\\ 
\end{tabular}
	\caption{\textbf{Mean joint position variance:} For each joint we compute the mean distance from every hypothesis to the mean prediction. In all cases the mean distance of the predictions for occluded joints is  higher than the one for visible joints. This can be used as a confidence measure. The head and neck joint were not regarded since less than 10 samples were occluded.}
	\label{tab:posevariance}
\end{table}

The Leeds Sports Pose dataset \cite{Johnson10} provides, together with the human pose annotations, the information whether a joint is visible or occluded.
We compute the mean distances of joint positions to the mean predicted skeleton for occluded and visible joints. Table \ref{tab:posevariance} shows that this variation is a good indicator for the uncertainty of the model as it is higher for occluded joints than for visible ones. Additionally, the variance for the end-effectors (hands, feet), which are the most difficult to predict, is much higher than for more stable points like hips and shoulders. 

\paragraph{Comparison to Mixture Density Networks}
Another way of dealing with uncertainty is explicitly estimating the density of the output distribution using MDNs \cite{bishop1994mixture}.
We note that MDNs differ from our method in two distinct points. 
First, MDNs estimate densities and our MHP model predicts multiple hypotheses instead. 
Second, MDNs are only well-defined for regression problems, whereas MHP models are agnostic to the loss and are thus more general.

We trained a MDN for human pose estimation. 
Although it is a relatively low dimensional problem (that is 14$\times$2D joints), it proved to be challenging for MDNs, especially since the Gaussians contain exponents with the number of dimensions ($c$ in Eq.~23 in \cite{bishop1994mixture}), which causes severe numerical problems.
In fact, we were unable to train the MDN with SGD with momentum, but had to resort to RMSProp as optimizer ([2] train with BGFS, a second order optimization technique, which is infeasible for deep networks due to the number of parameters). 
\begin{figure}[t]
	\centering
	\includegraphics[width=0.5\linewidth]{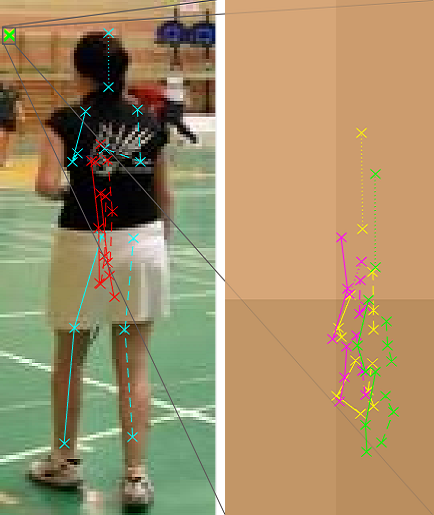}
	\caption{\textbf{MDNs for human pose estimation.} Mixture Density Networks become numerically unstable in higher dimensions, while at the same time suffer from degenerate predictions. The mixing coefficients for the degenerate predictions in the top left are almost 0, which lets all of their gradients vanish.}	
	\label{fig:mdn_hpe}
\end{figure}
In Figure \ref{fig:mdn_hpe} we compare the trained MDN with 5 Gaussians for the same image as the MHP cases in Figure \ref{fig:humanpose}.
The predicted probability for the blue skeleton is 98\%, $1.9\%$ for red and almost 0 for the remaining 3 (degenerated in top left corner).
The MDN is unable to recover more than one reasonable hypothesis, which is similar in every frame.
One reason is that all gradients for MDNs contain a multiplicative factor ($\alpha_i$ in \cite{bishop1994mixture}) for each component $i$ which prevents the model from learning mean and variance for this component once its $\alpha_i$ is close to 0. 

While MDNs have a clear advantage in predicting probabilities and variances together with the means, they are significantly more difficult to train and suffer from severe numerical instabilities in the high dimensional multivariate Gaussian distributions. Due to the simple nature of MHPs we are able to handle high dimensional problems without any stability issues. In the next section we address the task of future frame predition, for which we could not achieve convergence for MDNs.

\subsection{Future Frame Prediction}
\label{sec:frameprediction}
Predicting the future is inherently associated with ambiguity and as such, it is an ideal problem for multiple hypothesis prediction. 
The goal of future frame prediction is the pixel-wise estimation of a future frame in a video, given one or more previous frames, thus enclosing significant uncertainty. 
In this experiment we show that MHP models also extend to high dimensional problems, predicting images of resolution $128\times128\times3$ and $256\times256\times3$.
We use a fully convolutional residual architecture proposed by Laina et al.~\cite{laina2016deeper}, which has recently shown good potential for pixel-wise regression tasks, achieving state-of-the-art results on depth estimation without the need for additional refinement steps. 
We adapt the model to MHP, such that it predicts $M$ output maps with three channels each (RGB) by increasing the number of filters in the last up-sampling layer. 
All filters are initialized with ResNet-50 weights (pre-trained on ImageNet \cite{ILSVRC15} data), where possible, and random zero-mean Gaussian distributions with 0.01 standard deviation elsewhere.

\begin{figure}[!t]
	\centering
	\includegraphics[width=0.95\linewidth]{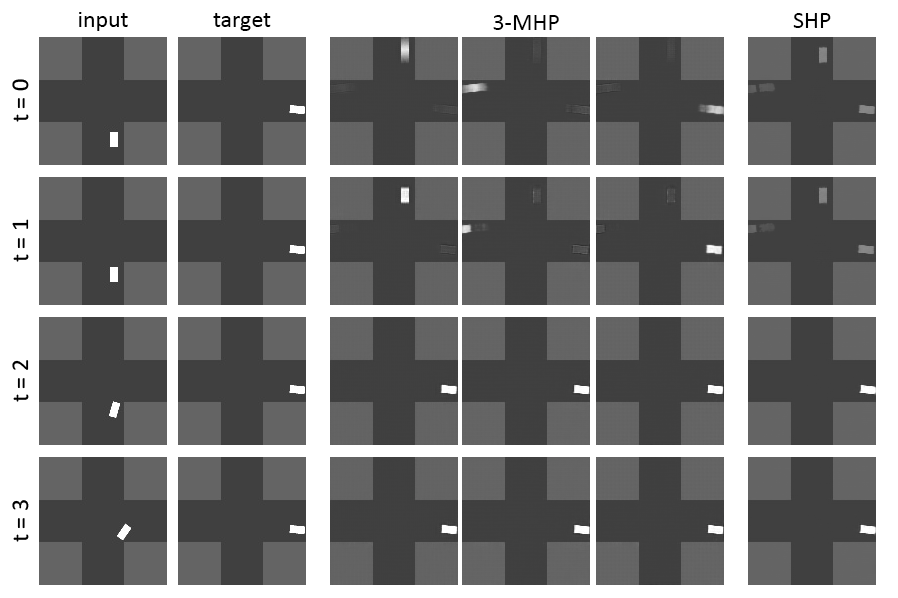}
	\caption{\textbf{Predicting the next frame on the synthetic Intersection dataset.} A SHP model is compared to a 3-MHP model, trained to predict the last frame of a sequence in which a car drives through an intersection. For $t={0,1}$, three outcomes are possible; SHP blurs them into one unrealistic frame with three ghost cars, whereas MHP predicts all three possible frames distinctly.
		\label{fig:intersection}}
	\vspace{-1.1em}
\end{figure}

\paragraph{Intersection}
The first dataset we use for future frame prediction is a simulation of a street intersection. 
We generate sequences where a simplified model car approaches the intersection from a random two-way road, slows down and then chooses one of the three possible routes to leave the crossing with equal probability. 
In this case, we are interested in predicting the last frame of the sequence, where the car is about to exit the view but still fully visible in the image. 
The dataset contains a discrete uncertainty regarding which exit the car will choose and a continuous uncertainty in the exact pose of the car in the last frame. 
We model this problem by training a network to predict three hypotheses about the future. 
Figure \ref{fig:intersection} shows a sample sequence. 
The first and second row show the single input frame and the target frame respectively. 
In the first two time stamps ($t={0,1}$), when the car is approaching the intersection and the destination is still unclear, the MHP outputs are distributed over the plausible outcomes as each hypothesis predicts a different possible exit location \ie north, east or west for the car coming from the south. The SHP model predicts an unrealistic frame where each exit shows a car which is the conditional average frame (see Equation \ref{eqn:conavg}).
At $t=2$ when the car starts taking a right turn, we observe that the three predictions collapse into a single decision (the eastern exit) with small variations in location and rotation to model the variance in exit pose. Here, the SHP model is also correct, since the uncertainty vanished.

The network is able to recognize whether a decision about the exit has been already made or not, and predicts a different selection of hypotheses in each of the two cases.
In the first two time steps, one can see faint ghost-cars for the non-selected exits; this is because of the balancing factor $\epsilon = 0.05$ that pulls the predictions slightly towards the conditional average, which is however necessary to avoid starving predictions during training, as detailed in Sec.~\ref{sec:minimizationscheme}. 

\paragraph{NTU Action Recognition Dataset}

Turning to real images, we evaluate the multiple hypothesis model on real data using the NTU RGB-D Action Recognition dataset~\cite{Shahroudy_2016_CVPR}. 
We use only the RGB videos for training and testing. 
Additionally, we automatically crop each sequence around the moving parts by thresholding the per pixel change between frames, since large parts of the frame are only static background.
The network is expected to learn the outcome of an action and predict the image at the end of the sequence. 
To analyze the image quality, we compute the mean gradient magnitude of a prediction, as a measure of sharpness:
\begin{equation}\label{eq:sharpness}
\mathcal{S}(f_\theta(x)) = \frac{1}{3whM}\sum_{c,p,j} ||G^j_c(p)||_2^2, \text{ where } G^j = \nabla f^j_\theta(x).
\end{equation}
$p$ iterates over pixel locations, $w$ and $h$ are the image dimensions and $c$ indexes the color channel.

\begin{figure}[t]
	\centering
	\includegraphics[width=0.95\linewidth]{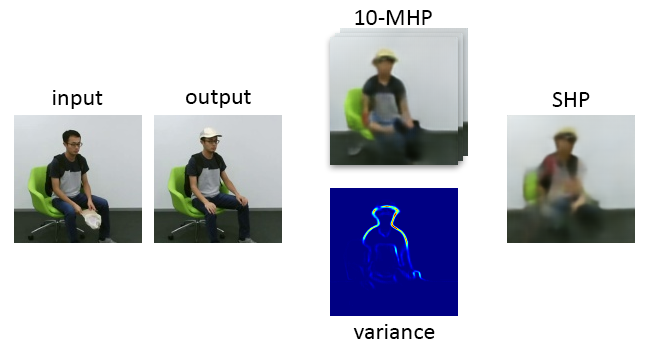}
	\caption{\textbf{Last-Frame Prediction.} Qualitative results for predicting the last frame of the \textit{put on a hat/cap} action. We show one randomly selected hypothesis. Again, SHP is very blurry, whereas MHP yields a sharper, distinct result. An additional benefit is the ability to compute per pixel variances over the predictions.}	
	\label{fig:hatprediction}
\end{figure}
\begin{table}[!t]
	\centering
	\small
	\begin{tabular}{ l | r | r }
		\textbf{Model} & \textbf{Sharpness} & \textbf{Min. MSE} \\ \hline
		SHP & 319.5 & 960.6 \\
		5-MHP & 359.2 & 808.2 \\
		10-MHP & \textbf{419.7} & \textbf{728.5} \\
	\end{tabular}
	\caption{\textbf{Sharpness and Error Analysis:} We measure the image sharpness (Eq.~\ref{eq:sharpness}, higher is better) for different numbers of hypotheses on the NTU dataset for the \textit{put on a hat/cap} action. Additionally, we report the average mean squared error (MSE) between the best prediction and the ground truth (lower is better).}
	\label{tab:sharpness}
\end{table}

In Table \ref{tab:sharpness} we compare the sharpness $\mathcal{S}$ for the \textit{put on a hat/cap} action.
With more predictions we produce sharper images and a lower error.
This effect can also be observed qualitatively in Figure \ref{fig:hatprediction}, where the improved image sharpness from 1 to 10 predictions becomes evident. 
Additionally, we display the per-pixel variance map which we compute in the case of multiple predictions. 
The map clearly identifies the person's head and shoulders as regions with higher estimated per-pixel uncertainty.
In this experiment we have shown that the MHP formulation extends to high-dimensional problems. Finally, we apply MHP to two discrete tasks: image segmentation and classification.

\begin{figure*}[ht]
	\centering
	\includegraphics[width=0.99\linewidth]{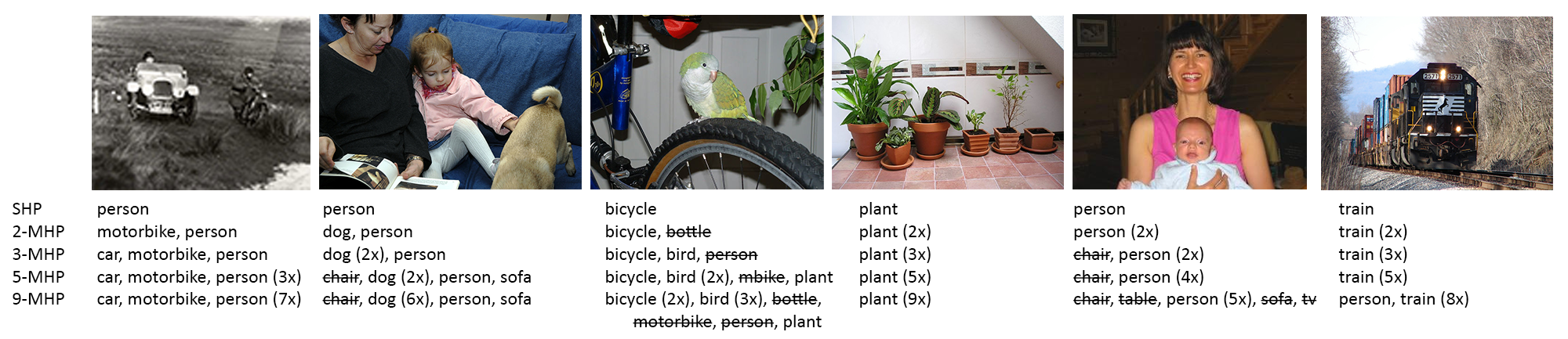}
	\caption{\textbf{Multiple Predictions on VOC 2012.} We show qualitative examples of multiple predictions. For each prediction we select the class with the maximum confidence. Networks with multiple predictions are able to identify several different classes in the images. The last image the ground truth annotation contains the \textit{person} label for the conductor in the train. Incorrect predictions are crossed out.}	
	\label{fig:vocclass}
\end{figure*}

\subsection{Multiple Object Classification}
\label{sec:classification}
Many previous approaches argue that single-label CNN models are not suitable for multi-label object recognition and propose multi-stage methods; we instead show that extending such a CNN architecture with the multiple hypothesis principle can achieve competitive performance for multiple labels, without the need for multi-stage pipelines.
We fine-tune a ResNet-101\footnote{ResNet-50 \cite{he2015deep} and VGG-16 \cite{Simonyan14c} behave similarly but with 2-3\% worse performance. For brevity we only show ResNet-101 results here.} pre-trained on ImageNet data and replace the output layer such that it predicts a set of $C$ class confidences for $M$ hypotheses ($C\cdot M$ values in total).

\begin{table}[t]
	\centering
	\small
	\begin{tabular}{ l | r | r | r | r }
		 & \textbf{VOC07} & \textbf{VOC12} & \textbf{COCO} & \textbf{COCO} \\ 
		\textbf{Method/Dataset} & mAP & mAP & mAP & mAP@10  \\ \hline
		WARP \cite{gong2013deep} & - & - & - & 49.2 \\
		HCP-1000 \cite{wei2014cnn} & 81.5 & - & - & -  \\ 
		CNN-RNN \cite{wang2016cnn} & 84.0 & - & - & 61.2  \\ \hline
		SHP (baseline) & 83.8 & 86.9 & 65.2 & 81.0  \\ \hline
		3-MHP (ours) & 84.1 & 87.3 & 66.1 & 82.2  \\
		5-MHP (ours) & 84.7 & 87.5 & \textbf{67.8} & \textbf{83.3}  \\
		9-MHP (ours) & \textbf{85.1} & \textbf{87.6} & 67.4 & 82.8 \\
		13-MHP (ours) & 84.7 & 87.0 & 67.7 & 83.1  \\
	\end{tabular}
	\caption{\textbf{Results on Pascal VOC 2007, 2012 and MS-COCO:} Classification results improve with more predictions over the single prediction baseline. At 9- and 13-MHP the performance decreases slightly due to false positives in some of the hypotheses as there are often much less true labels. (Results for \cite{gong2013deep} from \cite{wang2016cnn})
	}
	\label{tab:classification}
\end{table}

We can also address the problem of multi-label image classification as an MHP task, where $p(y|x)$ models the confidence that an instance of a certain class appears in the image $x$.
During training we give every image a probabilistic label that is uniformly selected from all classes that exist in the image.
For example, if an image contains two bikes and a person, every time the image is sampled during training it will be labeled either as \textit{bike} or \textit{person} with $50\%$ chance.
The network needs to resolve this label ambiguity. 

For evaluation, we use the 2007 and 2012 renditions of the Pascal Visual Object Classes (VOC)~\cite{Everingham10} dataset. There exist twenty different classes ($C=20$). 
In our experiments, we train the networks using the \emph{train} set of VOC2012 and evaluate their performance on the VOC2012 \emph{val} and VOC2007 \emph{test} splits. 
Additionally, we evaluate the MHP method on the MS Common Objects in Context (COCO) \cite{mscoco} containing $C=80$ classes, 82,783 training images and 40,504 validation images, which we use as testing data.
Here, the number of classes per image varies considerably.

In Table \ref{tab:classification} we show multi-label recognition results and compare them to three other methods using the mean average precision (mAP) and mAP@10 metrics.
mAP@$K$ computes the mAP for the $K$ classes that were detected with the highest confidence. We observe that all MHP models outperform the SHP baseline. In this discrete problem, it is natural that at high $M$ (in this case $\ge9$)  the performance decreases since there are often more predictions that possible discrete outcomes. In this case the additional hypotheses contribute some noise that reduces the scores slightly.

Figure \ref{fig:vocclass} shows qualitative results for different $M$.
We report the class with the highest confidence after soft-max of each prediction.
The networks trained with multiple predictions are able to identify additional objects in the image, as opposed to the single-label prediction.
When only a single class dominates the image, the predictions all tend to the same class.
For the qualitative results we use the class with the highest probability per hypothesis.

\subsection{Image Segmentation}
\label{sec:segmentation}
Finally, to be able to compare directly to multiple choice learning (MCL) \cite{lee2016stochastic} we trained a 4-MHP FCN8s \cite{long2015fully} for semantic segmentation on VOC2012. MCL trains separate networks making information exchange between ensemble members harder. Additionally, a full CNN needs to be trained for every single output of the ensemble, whereas adding more hypotheses does not add much overhead in our approach. Our model achieves a mean IoU of 70.3\%, compared to MCL's 69.1\% and uses 1/4 of the parameters (134.9M [ours] compared to 539.6M \cite{lee2016stochastic}).

In these last two experiments we showed that the MHP framework generalizes to discrete problems as well and thus is applicable for a wide variety of applications. 

	\section{Conclusions}
\label{conclusions}
We introduced a framework for multiple hypothesis prediction (MHP). This framework is principled, yielding a Voronoi tessellation in the output space, and simple, as it can easily be retrofitted to existing single hypothesis prediction (SHP) models and can be optimized with standard techniques such as backpropagation and gradient descent.

In an extensive set of experiments, we showed that MHP models routinely outperform their SHP counterparts, and that they simultaneously provide additional insights into the model. 
We demonstrated the representation of the output space as a Voronoi tessellation, the benefits of additional information in the variance over hypotheses and the applicability to high dimensional and discrete problems.
In future work, we hope to investigate the application of MHP models to time-series and other sequential data. 
\newline
\textbf{Acknowledgments}
This work was supported by the TUM - IAS (German Excellence Initiative - FP7 Grant 291763). 
	
	{\small
		\bibliographystyle{ieee}
		\bibliography{bibliography}
	}
	
\end{document}